\documentclass[letterpaper, 10 pt, conference]{ieeeconf}  

\IEEEoverridecommandlockouts                              

\usepackage{cite}
\usepackage{amsmath,amssymb,amsfonts}
\usepackage{algorithmic}
\usepackage{graphicx}
\usepackage{textcomp}
\usepackage{xcolor}
\usepackage{comment}
\usepackage[style=base]{caption}
\usepackage{subcaption}
\usepackage{pdfpages}
\usepackage{amsmath}
\usepackage{bm}
\usepackage{siunitx}
\usepackage{placeins}
\usepackage{mathtools}
\usepackage{xfrac}
\usepackage{url}
\usepackage{gensymb}
\usepackage{footmisc}
\usepackage{tabularx}
\usepackage[colorlinks=true, allcolors=black]{hyperref}
\usepackage{cleveref}

\DeclarePairedDelimiter\ceil{\lceil}{\rceil}
\newcommand{\ts}{\textsuperscript}

\title{\LARGE \bf
Towards High-Definition Maps: a Framework Leveraging Semantic Segmentation to Improve NDT Map Compression and Descriptivity
}

\author{Petri Manninen$^{1}$, Heikki Hyyti$^{1}$, Ville Kyrki$^{2}$, Jyri Maanpää$^{1}$, Josef Taher$^{1}$ and Juha Hyyppä$^{1}$
\thanks{*This work was supported by Academy of Finland, decisions 337656, 319011, 318437 and by Henry Ford foundation Finland.}
\thanks{$^{1}$P. Manninen, H. Hyyti, J. Maanpää, J. Taher, J. Hyyppä are with Department of Remote Sensing and Photogrammetry,
              Finnish Geospatial Research Institute (FGI), National Land Survey of Finland (NLS),
              02150 Espoo, Finland {\tt\small petri.manninen@nls.fi, heikki.hyyti@nls.fi, jyri.maanpaa@nls.fi, josef.taher@nls.fi, juha.hyyppa@nls.fi}}%
\thanks{$^{2}$V. Kyrki is with School of Electrical Engineering, Aalto University, 02150 Espoo, Finland {\tt\small ville.kyrki@aalto.fi}}%
}

\usepackage{fancyhdr}
\fancypagestyle{firststyle}
{
\fancyhf{} 
\fancyfoot[LO]{\scriptsize \textcopyright 2022 IEEE. Personal use of this material is permitted.
  Permission from IEEE must be obtained for all other uses, in any current or future
  media, including reprinting/republishing this material for advertising or promotional
  purposes, creating new collective works, for resale or redistribution to servers or
  lists, or reuse of any copyrighted component of this work in other works.
  DOI: \href{https://doi.org/10.1109/IROS47612.2022.9982050}{10.1109/IROS47612.2022.9982050}}
}

\begin{document}

\maketitle
\thispagestyle{firststyle}
\pagestyle{empty}

\begin{abstract}

High-Definition (HD) maps are needed for robust navigation of autonomous vehicles, limited by the on-board storage capacity. To solve this, we propose a novel framework, Environment-Aware Normal Distributions Transform (EA-NDT), that significantly improves compression of standard NDT map representation. The compressed representation of EA-NDT is based on semantic-aided clustering of point clouds resulting in more optimal cells compared to grid cells of standard NDT. To evaluate EA-NDT, we present an open-source implementation that extracts planar and cylindrical primitive features from a point cloud and further divides them into smaller cells to represent the data as an EA-NDT HD map. We collected an open suburban environment dataset and evaluated EA-NDT HD map representation against the standard NDT representation. Compared to the standard NDT, EA-NDT achieved consistently at least 1.5$\bm{\times}$ higher map compression while maintaining the same descriptive capability. Moreover, we showed that EA-NDT is capable of producing maps with significantly higher descriptivity score when using the same number of cells than the standard NDT.
\end{abstract}

\section{INTRODUCTION} \label{sec:introduction}

The current development of mobile robots and the on-going competition for the crown of autonomous driving has increased the demand of accurate positioning services. Generally, Global Navigation Satellite System (GNSS) can be used to measure global position of a mobile robot but the accuracy of satellite navigation alone is typically around a few meters and because of signal obstruction the satellite signals may be unavailable \cite{zaidi2006global, wang2002pseudolite}. Alternatively, global position can be solved by fitting the current sensor view into an existing georeferenced map, that can be computed e.g. with Simultaneous Localization and Mapping (SLAM) \cite{cadena2016past}. Moreover, map-based technique provides a combined position and rotation estimate in contrast to a global position measured by GNSS. 

Maps used in autonomous driving are typically called High-Definition (HD) maps \cite{seif2016autonomous, liu2020high}. Data compression of HD maps is of high importance within many applications that have limited computational resources and storage capacity \cite{dube2020segmap, yin20203d, chang2021map}. Moreover, real-time localization requires compressed maps to ensure fast processing capability.

Since positioning in real-time with raw point clouds is infeasible, alternative methods have been developed to overcome the problem \cite{magnusson2009three, stoyanov2012fast, zaganidis2017semantic, zaganidis2018integrating, dube2020segmap}. One promising approach is Normal Distributions transform (NDT) \cite{magnusson2009three}. NDT compresses the three-dimensional point cloud data by dividing the cloud into equal sized cubical cells that are expressed by their mean and covariance. To improve the scan registration, Semantic-assisted Normal Distributions Transform (SE-NDT) \cite{zaganidis2018integrating} expanded the original NDT with semantic information. However, both NDT and SE-NDT use a grid structure for the division of the point cloud, and therefore cannot find the fundamental geometrical structure of the environment (e.g. boundaries between object surfaces). Consequently, this results in an NDT representation where part of the cells have a high variance in all three dimensions. Magnusson \cite{magnusson2009three} also presented that the point cloud can alternatively be divided by K-means clustering \cite{sammut2017encyclopedia} and that it improves the scan registration compared to using a grid structure.

\begin{figure}[!t]
    \centering
    \includegraphics[width=\linewidth]{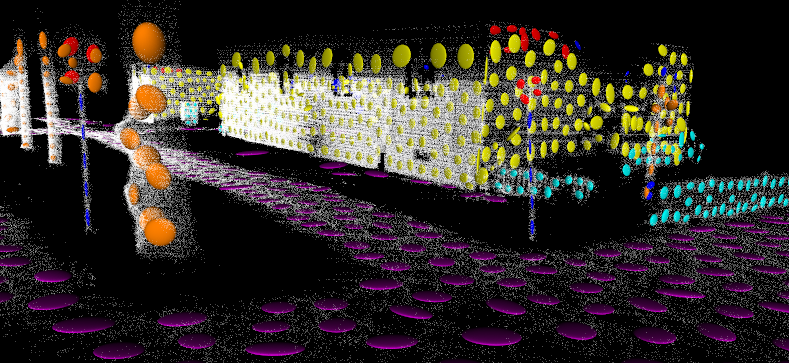}
    \caption{An illustration of a point cloud (white) and corresponding EA-NDT HD map representation. EA-NDT cells are visualized with ellipsoids (mass within a standard deviation) presenting building (yellow), fence (cyan), ground (purple), pole (blue), tree trunk (orange) and traffic sign (red) labels.}
    \label{fig:visual abstract}
\end{figure}

In this work, we address the aforementioned problem of sub-optimal point cloud division. We propose to solve the problem by leveraging semantic-aided clustering. We present a novel framework called Environment-Aware NDT (EA-NDT) (illustrated in Fig.~\ref{fig:visual abstract}), which provides EA-NDT HD Map representation. EA-NDT HD Map is a compressed representation of a point cloud that is based on the leaf cell representation of the standard NDT, and therefore NDT scan registration technique is directly applicable with EA-NDT. In this work, the standard NDT is referred as NDT. In contrast to the grid structure of NDT, EA-NDT leverages semantic information to cluster planar and cylindrical primitives of the environment to provide a more optimal NDT cell division. In EA-NDT HD Map, each cell only consists of points that model the same basic geometrical shape such as a plane or a pole. Moreover, by adding understanding of semantic information in the scene, we can compute a map containing only stable objects that are useful for accurate localization.

The main contributions of this paper are: 
\begin{enumerate}
    \item A novel data-driven framework to compute NDT map representation without the grid structure.
    \item Demonstration of significantly improved data compression compared to NDT representation.
    \item An open-source implementation\footnote{\url{https://gitlab.com/fgi_nls/public/hd-map}\label{fn1}} of the proposed EA-NDT is shared for the community.
    \item A registered dataset\footnote{\url{https://doi.org/10.5281/zenodo.6796874}\label{fn2}} to evaluate the proposed EA-NDT on data collected with Velodyne VLS-128 LiDAR.
\end{enumerate}

The rest of the paper is organized as follows: The next section describes the related work in the fields of HD Maps, scan registration, SLAM, and point cloud semantic segmentation. In Section~\ref{sec:framework}, we formalize a pipeline architecture of the proposed framework to extract planar and cylindrical primitives of the point cloud. The implementation details of our proof of concept solution are explained in Section~\ref{sec:experiments} together with an introduction to the data collection setup and preprocessed dataset, and the evaluation metrics used for the experiment. In Section~\ref{sec:results}, we compare the proposed EA-NDT to a map computed with NDT and show that EA-NDT provides a significant map compression while maintaining the same descriptive capability. Finally, in Section~\ref{sec:discussion} we consider the advantages and disadvantages of EA-NDT and provide a discussion over the validity, reliability and generalizability of the experiment.

\section{RELATED WORK} \label{sec:related work}

HD Maps are one of the key techniques to enable autonomous driving \cite{liu2020high}. Seif and Hu \cite{seif2016autonomous} have recognized three challenges to be solved with an HD map: the localization of the vehicle, reacting to events beyond sight, and driving according to the needs of the traffic. In this work, we focus in the localization task. An HD Map can be computed e.g. with SLAM that is a well established and profoundly studied problem about how to align subsequent sensor measurements to incrementally compute a map of the surrounding environment while simultaneously localizing the sensor \cite{bresson2017simultaneous}. In the review by Bresson et al \cite{bresson2017simultaneous}, they found that an accuracy of 10 cm has been reported for the built maps but even an accuracy of 2 cm is possible.

Data compression is a crucial challenge for HD maps in large environments. For example, the well known Iterative Closest Point (ICP) \cite{besl1992icp} algorithm is infeasible in large point clouds due to the computational cost of finding closest corresponding points across a measurement and a map. To improve the computational problems of ICP, Magnusson proposed Point-to-Distribution (P2D)-NDT in which the reference cloud is divided by a fixed sized 3D grid into cells modelled by the mean and covariance of the points \cite{magnusson2009three}. In (P2D)-NDT, each point in the registered scan is fitted to the cells within a local neighbourhood of the point. In addition to robust scan registration, NDT representation provides data compression together with faster registration. Stoyanov et al. presented Distribution-to-Distribution (D2D)-NDT that further develops the P2D-NDT to likewise model the registered scan with normal distributions \cite{stoyanov2012fast}.

Semantic information can enhance scan registration performance of NDT. For example, Semantic-assisted NDT (SE-NDT) \cite{zaganidis2017semantic}, proposed by Zaganidis et al., showed that the use of even two semantic labels (edges and planes) can improve the scan registration. To further develop SE-NDT, Zaganidis et al. presented a complete semantic registration pipeline that uses a deep neural network for semantic segmentation of the point cloud \cite{zaganidis2018integrating}. SE-NDT uses the 3D grid cell structure of NDT but models each semantic label separately to utilize the division of similar entities in the registration task. For semantic segmentation, SE-NDT uses PointNet \cite{qi2017pointnet} that is a pioneering solution of a point cloud segmentation network that consumes raw point cloud data without voxelization or rendering. Cho et al. proposed that the uncertainty of semantic information could also be used in the registration task \cite{cho2020semantic}.

Semantic information can be utilized further than was proposed in the previous works. In this work, we propose to replace the aforementioned grid division by leveraging semantic-aided clustering that finds planar and cylindrical structures of a point cloud. For semantic segmentation, we use Random sampling and an effective Local feature Aggregator-Net (RandLA-Net) \cite{hu2020randla} that presents a new local feature aggregation module to support random sampling that was found a suitable technique for semantic segmentation of large scale point clouds. They have reported up to 200$\times$ faster processing capacity compared to the existing solutions. A further review of semantic segmentation of point cloud data is available e.g. in \cite{zhang2019review}.

\section{ENVIRONMENT-AWARE NDT} \label{sec:framework}

Here we propose a framework, called Environment-Aware NDT (EA-NDT), to divide a semantically segmented point cloud into NDT cells. The proposed framework is a straight pipeline process consisting of 4 stages (Fig.~\ref{fig:HD_map_pipeline}) that step-by-step divide the input point cloud into cells which are ultimately represented as an NDT map. The input of the pipeline is a \textbf{Registered Point Cloud}, which is processed in the following order by stages called \textbf{Semantic Segmentation}, \textbf{Instance Clustering}, \textbf{Primitive Extraction} and \textbf{Cell Clustering}. Finally, the output of the pipeline is an environment-aware NDT-based HD map representation, called \textbf{EA-NDT HD Map}, which stores the found cells using NDT representation. 

In the \textbf{Registered Point Cloud} each 3D point has X, Y, Z Cartesian coordinate (e.g. ETRS-TM35FIN, ECEF) and an intensity value. \textbf{Semantic Segmentation} appends semantic information for each point in the cloud to enable further clustering of the data. In this work, we used road, sidewalk, parking, building, fence, pole, traffic sign, and tree trunk labels to demonstrate the framework but also other labels could be used. \textbf{Instance Clustering} divides each semantic segment into instances that are spatially separated from each other. \textbf{Primitive Extraction} divides each instance into predefined primitives that can be modeled with an unimodal distribution. In this work, we have defined planar and cylindrical primitives but the framework could be extended to support new types of primitives. However, large primitives such as trees can not be modelled well with an uniform distribution. Therefore, \textbf{Cell Clustering} further divides each primitive into cells of approximately equal size while minimizing the number of used cells. Ultimately, \textbf{EA-NDT HD Map} is presented as an octree \cite{kammerl2012real} that in this work stores the point counter, point sum and upper diagonal of the covariance matrix for each cell but other attributes such as semantic segment, instance cluster or primitive type could be included.

\section{METHODS AND EXPERIMENTS} \label{sec:experiments}

To demonstrate the proposed framework to build an \textbf{EA-NDT HD Map}, we used a dataset collected with Velodyne VLS-128 Alpha Puck \cite{Velodyne:VLS128} LiDAR 7th of September 2020 in a suburban environment in the area of Käpylä in Helsinki, the capital of Finland. The environment in the dataset consists of a straight two-way asphalt street, called Pohjolankatu, which starts from a larger controlled intersection at the crossing of Tuusulanväylä (60.213326$\degree$ N, 24.942908$\degree$ E in WGS84) and passes by three smaller uncontrolled intersections until the crossing of Metsolantie (60.215537$\degree$ N, 24.950065$\degree$ E). It is a typical suburban street with tram lines, sidewalks, small buildings, traffic signs, light poles, and cars parked on both sides of the streets. To collect a reference trajectory and to synchronize the LiDAR measurements, we have used a Novatel PwrPak7-E1 GNSS Inertial Navigation System (INS) \cite{Novatel:PwrPak7-E1}. The sensors were installed on a Ford Mondeo Hybrid research platform named Autonomous Research Vehicle Observatory (ARVO) \cite{maanpaa2021multimodal}. The sensors were interfaced through Robotic Operation System (ROS) \cite{quigley2009ros} version Kinetic Kame and the sensor measurements were saved in rosbag format for further processing.

\subsection{Preprocessed Dataset} \label{subsec:experiments:dataset}

\begin{figure}[!b]
    \centering
    \includegraphics[width=\linewidth]{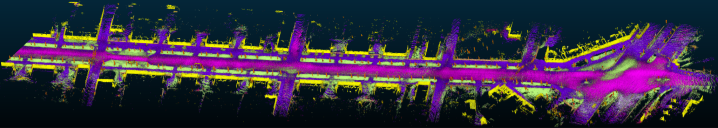}
    \caption{The complete dataset visualized with semantic labels in different colors: road (magenta), sidewalk (violet), parking (pink), terrain (green), buildings (yellow), fence (light brown), tree trunk (brown), traffic sign (red), pole (grey).}
    \label{fig:dataset}
\end{figure}

Our open preprocessed dataset\footref{fn2}, shown in Fig.~\ref{fig:dataset}, consists of a two-way asphalt paved street with a tram line to both directions and sidewalks in both sides of the street. The length of the dataset trajectory is around 640 m and it has in total more than 40 million points from which 28 million are used in this work. All the intersections together have a plenty of traffic signs. The sidewalks are separated from the road by a row of tall planted trees. The dataset contains nearly 30 buildings that are mostly wooden and there are several fences between the houses. Our dataset includes all the semantic labels classified by RandLA-Net \cite{hu2020randla} but in this work we have used only road, sidewalk, parking, building, fence, tree trunk, traffic sign, and pole labels. In this work, road, sidewalk, and parking labels were reassigned into a common ground label. The proportion and the number of the points of each label are shown in Table~\ref{table:dataset_characteristics}. Half of the used points consist of ground and roughly a fourth represent buildings whereas poles and traffic signs together represent only 1 \%. Tree trunks and fences together represent a fifth of the used points. The preprocessing of the data consists of three steps; semantic segmentation, scan registration, and data filtering. 

In semantic segmentation of scans, we used a RandLA-Net model pre-trained with SemanticKITTI dataset which was collected with Velodyne HDL-64 LiDAR \cite{behley2019semantickitti}. Instead, we used VLS-128 which has a longer range and 128 laser beams instead of 64 \cite{Velodyne:VLS128}. Also, VLS-128 has a wider field of view (FOV) in vertical direction. Consequently, the measurements outside of the vertical FOV of HDL-64 were constantly misclassified so only the measurements within HDL-64 FOV were used. RandLA-Net outputs a probability estimate vector of labels for each point. In this work, we call it as label probabilities.

\begin{table}[!t]
\renewcommand{\arraystretch}{1.3}
\sisetup{round-mode=places}
\centering
\caption{RandLA-Net classified dataset label proportions.}
\label{table:dataset_characteristics}
\begin{tabular}{ l  r  S[round-precision=1,table-align-text-post=false]  S[round-precision=1,table-align-text-post=false] }
    \bfseries Semantic label & \bfseries No. of points & \bfseries {\% of all} & \bfseries {\% of used} \\ 
    \hline\hline
    Ground                              & 14,052,836    & 34.7312265 & 50.8023727 \\
    Building                            & 7,650,980     & 18.9092024 & 27.6590389 \\
    Tree trunk                          & 3,560,910     &  8.8006984 & 12.8730369 \\
    Fence                               & 2,120,849     &  5.2416243 &  7.667076  \\
    Pole                                & 193,516       &  0.4782699 &  0.6995792 \\
    Traffic sign                        & 82,680        &  0.2043415 &  0.2988963 \\
    \hline
    Labels used here                    & 27,661,771    & 68.365363 &  100.0 \\
    Others                              & 12,799,904    & 31.634637 &   \\
    \hline
    Total                               & 40,461,675    & 100.0 &  \\
\end{tabular}
\end{table}

In scan registration, the motion deformation of each scan was first fixed according to a GNSS INS trajectory post-processed with Novatel Inertial Explorer \cite{Novatel:Inertial_explorer}, after which P2D-NDT implementation \cite{koide3:ndt_omp} with 1 m grid cell size was used for registration. In registration, a local map of 5 last keyframes was used as a target cloud and a motion threshold of 10 cm was used to add a new keyframe. Grid cells containing points from a single ring of the LiDAR were ignored in the registration. Moreover, points that were considered possibly unreliable (vehicles, bicycles, pedestrians and vegetation) or further than 50 m away from the LiDAR, were ignored.

\begin{figure*}[!t]
    \centering
    \includegraphics[width=\textwidth]{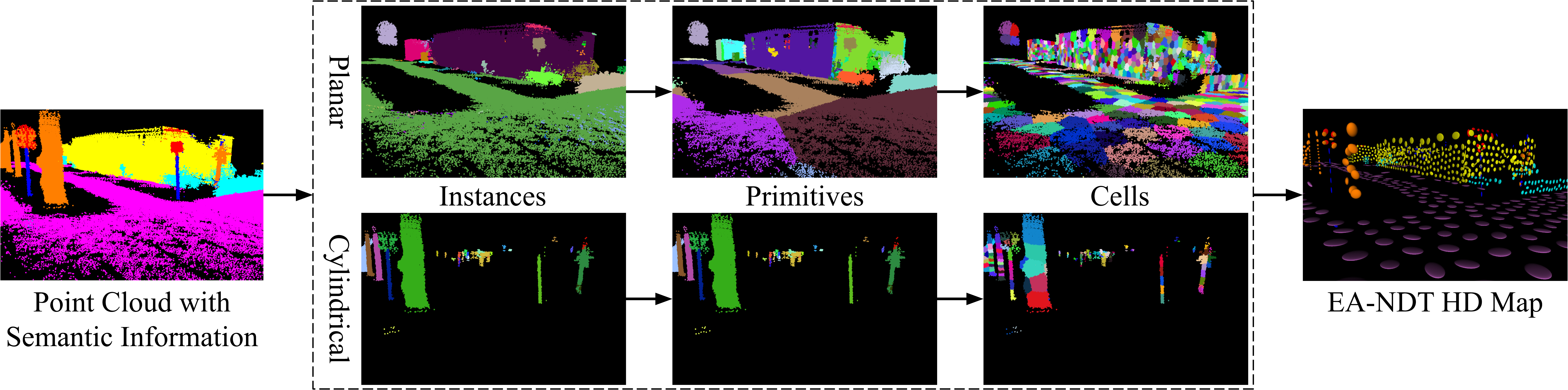}
    \caption{A visualization of the proposed EA-NDT processing pipeline that is based on the framework in Section~\ref{sec:framework}. The input is a semantically segmented point cloud and the intermediate phases before EA-NDT HD Map are instances, primitives and cells, in which the entities are separated by color. The color mapping of semantic information is explained in Fig.~\ref{fig:visual abstract}.}
    \label{fig:HD_map_pipeline}
\end{figure*}

After the scan registration, the dense \textbf{Registered Point Cloud} was voxel filtered to average X, Y, Z position and label probabilities of each 1 cm voxel. To smoothen the semantic segmentation, the label probabilities of each point was averaged within a radius of 5 cm.

\subsection{The Implementation} \label{subsec:experiments:implementation}

A method\footref{fn1}, based on the framework presented in Section~\ref{sec:framework}, was implemented in C++14 on top of ROS Noetic Ninjemys. The main functionality of the implementation uses existing functions and classes of Point Cloud Library (PCL) \cite{Rusu_ICRA2011_PCL}. The implemented processing pipeline is demonstrated in Fig.~\ref{fig:HD_map_pipeline}. The 1\ts{st} stage of the framework, \textbf{Semantic Segmentation}, is explained in Section~\ref{subsec:experiments:dataset}. Therefore, our dataset already includes the semantic information.

\textbf{Instance clustering} was implemented with Euclidean region growing algorithm \cite{trevor2013efficient} to divide each semantic segment into spatially separate instances shown in Fig.~\ref{fig:HD_map_pipeline}. In general, we require a distance threshold of 30 cm between the instances and a minimum of 10 points per instance. For ground label we require a distance threshold of 50 cm between the instances and a minimum of 3000 points per instance. In our dataset, there is a significant amount of outliers and reflected points below the ground plane that are undesired in a map, \textbf{Instance clustering} is used to filter those points.

In \textbf{Primitive Extraction}, tree trunk and pole instances are modeled as individual cylindrical primitives, and traffic sign instances as individual planar primitives. Both primitive types are shown in Fig.~\ref{fig:HD_map_pipeline}. For other semantic labels, planar primitives were extracted by Random Sample Consensus (RANSAC) \cite{fischler1981random} based normal plane fitting algorithm after subsampling the instance with an averaging 10 cm voxel grid and estimating the point normals for each remaining point from 26 nearest neighbours. For building and fence instances, a normal distance weight of $\sfrac{\pi}{4}$ and a distance threshold of 15 cm was used for plane fitting. For ground instances, the procedure differs slightly: 1) an existing implementation \cite{genbattle:dkm} of K-means++ algorithm \cite{arthur2007k} was used to divide the ground instances into primitives with an area of approximately 100 m² (The initialization of number of K-means clusters is explained later in this section in \textbf{Cell Clustering}), after which 2) the plane fitting was performed for each primitive with a 30 cm distance threshold for a coarse noise filtering.

\begin{table}[!b]
\renewcommand{\arraystretch}{1.3}
\sisetup{round-mode=places}
\centering
\caption{The final values of the scaling parameters}
\label{table:tuning_parameters}
\begin{tabular}{  l  S[round-precision=3,table-align-text-post=false]  S[round-precision=3,table-align-text-post=false] }
    \bfseries Semantic label ($L$) & {\bfseries $f_L$} & {\bfseries $g_L$} \\ 
    \hline\hline
    Ground                              & 1.6803924146591254 &  0.08305231866698243 \\
    Building                            & 2.7078758377808536 &  0.13722034139500836 \\
    Tree trunk                          & 4.17948650640806   &  0.31843996435228533 \\
    Fence                               & 2.2479127883095584 & -0.7883008443523578  \\
    Pole                                & 1.6874096382321715 & -0.31506643695059683 \\
    Traffic sign                        & 3.9231919696267386 &  0.3165458127096211  \\
\end{tabular}
\end{table}

\begin{figure}[!b]
    \centering
    \includegraphics[width=\linewidth]{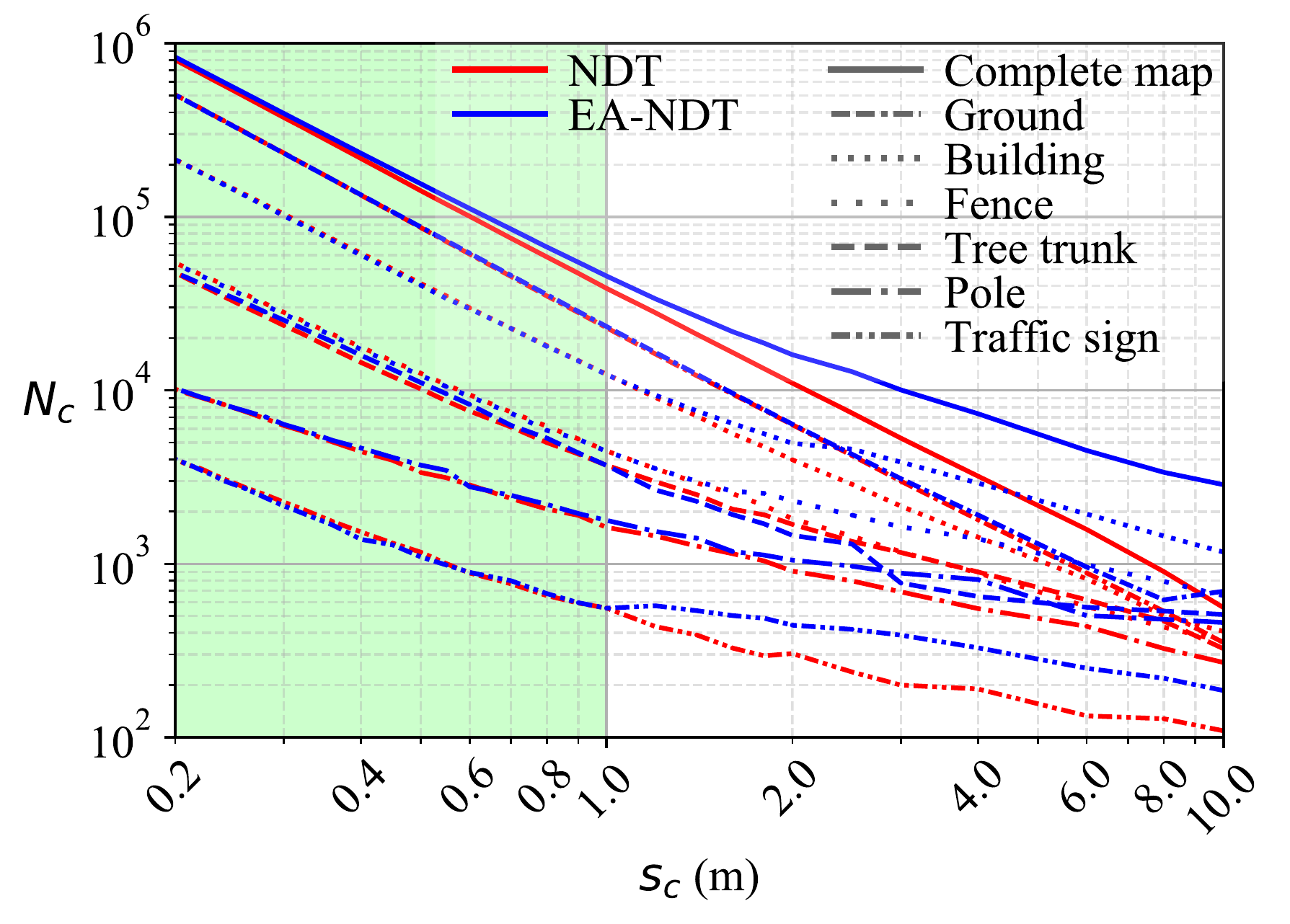}
    \caption{The number of cells $N_c$ after fitting EA-NDT with NDT shown w.r.t. cell size $s_c$, color indicates the method, line style the label, and green background the fitted range.}
    \label{fig:fitting_leaves_and_resolutions}
\end{figure}

In \textbf{Cell Clustering}, primitives are divided into cells (shown in Fig.~\ref{fig:HD_map_pipeline}) with K-means++ algorithm for which the number of clusters
\begin{equation}
    N_L = \ceil*{f_L {n_L}^{g_L}}
    \label{eq:fit_num_cells}
\end{equation}
is initialized for each label $L$. In \eqref{eq:fit_num_cells}, $\ceil*{\cdot}$ is the ceiling operator, $n_L$ is either $n_{\alpha}$ for cylindrical primitives (tree trunk and pole) or $n_{\beta}$ for planar primitives (ground, building, fence, and traffic sign): 
\begin{equation}
    n_{\alpha} = {l_{\alpha}} \big/ {s_c}
    \qquad
    \text{and}
    \qquad
    n_{\beta} = {A_{\beta}} \big/ {{s_c}^2},
    \label{eq:cluster_num}
\end{equation}
where $l_{\alpha}$ is the length of a cylindrical primitive and $A_{\beta}$ is the number of points remaining after projecting the planar primitive into the eigenspace found by principal component analysis (PCA) and filtering with a 10 cm voxel grid. Additionally, after clustering cells in ground, a plane fitting with a 15 cm threshold is performed for each cell for a finer noise filtering.
In \eqref{eq:fit_num_cells}, scaling parameters $f_L$ and $g_L$, shown in Table~\ref{table:tuning_parameters}, were manually fitted for each $L$ over 6 iterations starting from $f_{L_0}=1$ and $g_{L_0}=1$ until the number of cells $N_c$ for EA-NDT (shown in Fig.~\ref{fig:fitting_leaves_and_resolutions}) was sufficiently close to NDT with cell size $s_c < 1\:\textrm{m}$. Despite the cell size, each primitive is required to have at least one cell. Fig.~\ref{fig:fitting_leaves_and_resolutions} reveals how this sets a lower boundary for $N_c$ with larger cells.

Finally, all the computed cells were stored into an octree structure \cite{kammerl2012real} that represents the \textbf{EA-NDT HD Map}. Each leaf cell in the octree stores a point counter, point sum and upper diagonal of the covariance matrix for the cell. We require a minimum of 6 points for a leaf cell to be modeled reliably with a normal distribution, hence cells with less points are ignored. The octree implementation in PCL requires a minimum leaf cell size parameter, we used $\sfrac{s_c}{4}$ to make it sufficiently smaller compared to the required cell size of EA-NDT.

\begin{figure}[!t]
    \centering
    \includegraphics[width=\linewidth]{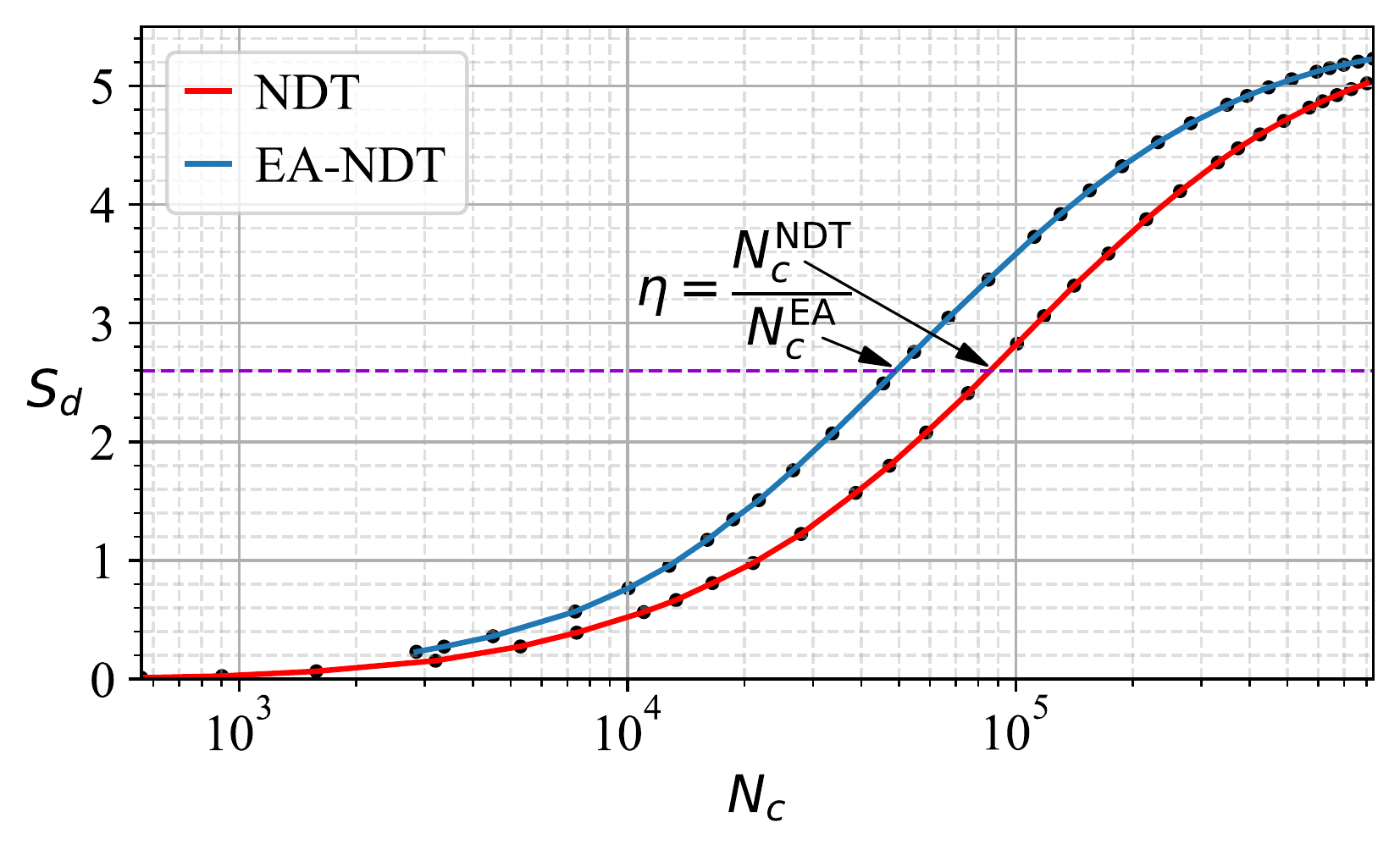}
    \caption{The complete map descriptivity score $S_d$ compared w.r.t. number of cells $N_c$. The violet line depicts the computation of the NDT compression efficiency $\eta$ for each $S_d$.}
    \label{fig:complete score vs number of leafs}
\end{figure}

\subsection{Evaluation} \label{subsec:experiments:evaluation}

Here, a descriptivity score $S_d$, in which a higher score denotes higher similarity, is defined to evaluate how well the map models the raw point cloud. It is derived using a density function of a multivariate normal distribution \cite{prince2012computer}, which is defined for each 3D point $\mathbf{x}_i$ and $j$\ts{th} NDT cell as
\begin{equation}
    f_j(\mathbf{x}_i) = \frac{1}{\sqrt{(2\pi)^k|\mathbf{\Sigma}_j}|} 
    e^{ \left(-\frac{1}{2}(\mathbf{x}_i-\boldsymbol{\mu}_j)^\mathsf{T}\mathbf{\Sigma}_j^{-1}(\mathbf{x}_i-\boldsymbol{\mu}_j) \right)}.
    \label{eq:density_function}
\end{equation}
In \eqref{eq:density_function}, $\boldsymbol{\mu}_j$ is the mean vector of a distribution with a covariance matrix $\boldsymbol{\Sigma}_j$ for $j$\ts{th} NDT cell, ${|\cdot|}$ is the determinant operator and $k=3$ describes dimension of the multivariate distribution. 
Restricting to the local neighborhood of each $N_p$ point, the descriptivity score $S_d$ is an average density of best fitting NDT cells:
\begin{equation}
    S_d = \frac{1}{N_p} \sum_{i=1}^{N_p} \ \mathop{\max f_j(\mathbf{x}_i)}_{{\lVert\mathbf{x_i}-\boldsymbol{\mu}_j\rVert}_2 \, \leq \, 2 s_c}.
    \label{eq:average_score}
\end{equation}
The maximum distance inside a grid cell is $\sqrt{3} {s_c}$, and therefore the radius of $2 s_c$ was considered large enough to contain the highest fit. 

\begin{figure}[!t]
    \centering
    \includegraphics[width=\linewidth]{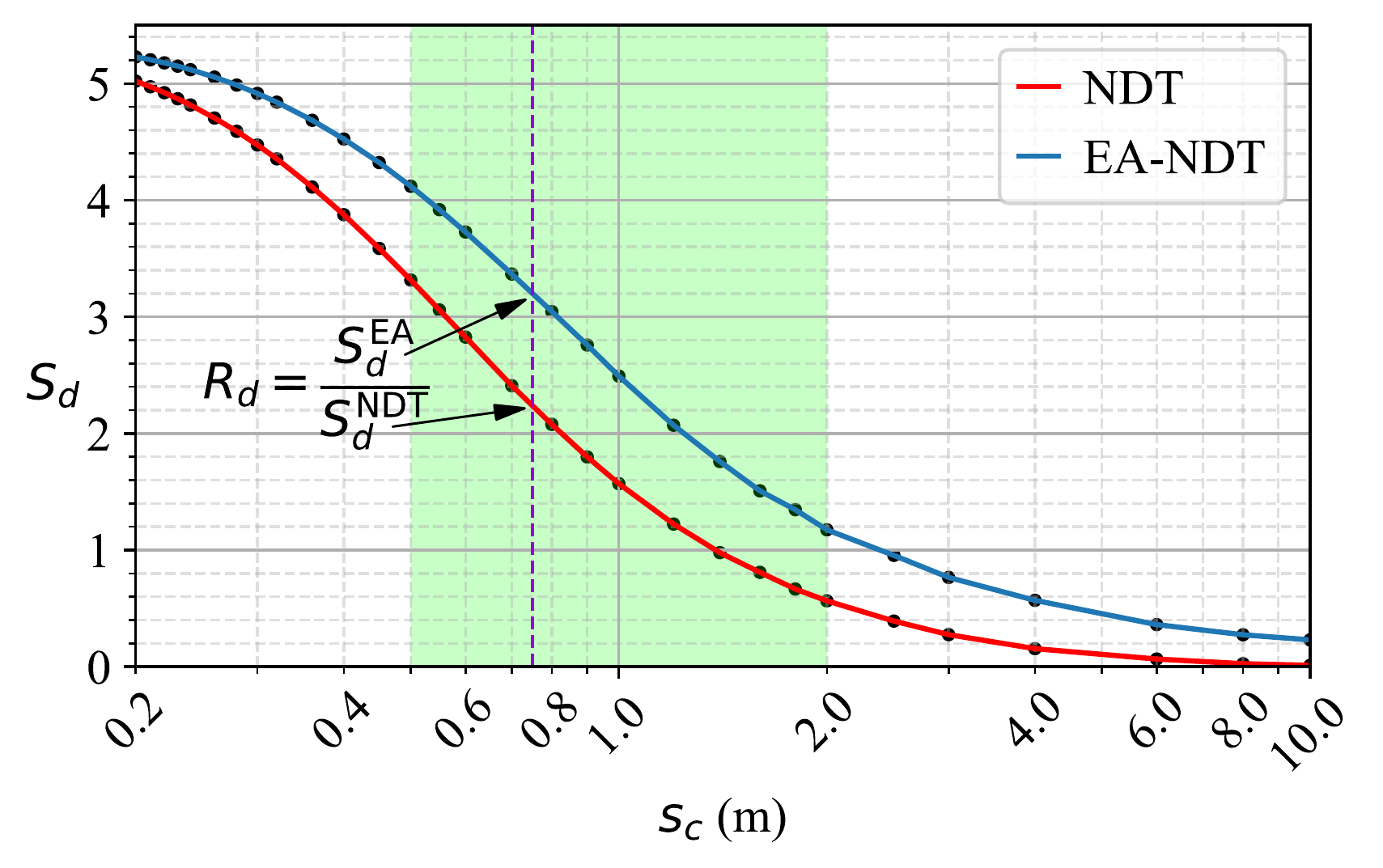}
    \caption{An alternative comparison of the complete map descriptivity score $S_d$ w.r.t. cell size $s_c$. The violet line depicts the computation of the descriptivity ratio $R_d$ for each $s_c$ and green background emphasizes the applicable range.}
    \label{fig:complete score vs resolution}
\end{figure}

We have defined two ratios, descriptivity ratio $R_d$ requiring $s^\text{EA}_c=s^\text{NDT}_c$ (Fig.~\ref{fig:complete score vs resolution}) and data compression ratio $R_c$:
\begin{equation}
    R_d = {S^\text{EA}_d} \big/ {S^\text{NDT}_d}
    \qquad
    \text{and}
    \qquad
    R_c = {N_p \sigma_p} \big/ {N_c \sigma_c},
    \label{eq:ratios}
\end{equation}
where superscripts EA and NDT stand for EA-NDT and NDT, respectively, $\sigma_p$ and $\sigma_c$ are the data size of the point and the cell, respectively. Using \eqref{eq:ratios} while requiring $S^\text{EA}_d=S^\text{NDT}_d$, we define an NDT compression efficiency (Fig.~\ref{fig:complete score vs number of leafs})
\begin{equation}
    \eta = {R^\text{NDT}_c} \big/ {R^\text{EA}_c} = {N^\text{NDT}_c} \big/ {N^\text{EA}_c}.
    \label{eq:ratio_of_compression}
\end{equation}

\section{Results} \label{sec:results}

\begin{figure}[!t]
    \centering
    \includegraphics[width=\linewidth]{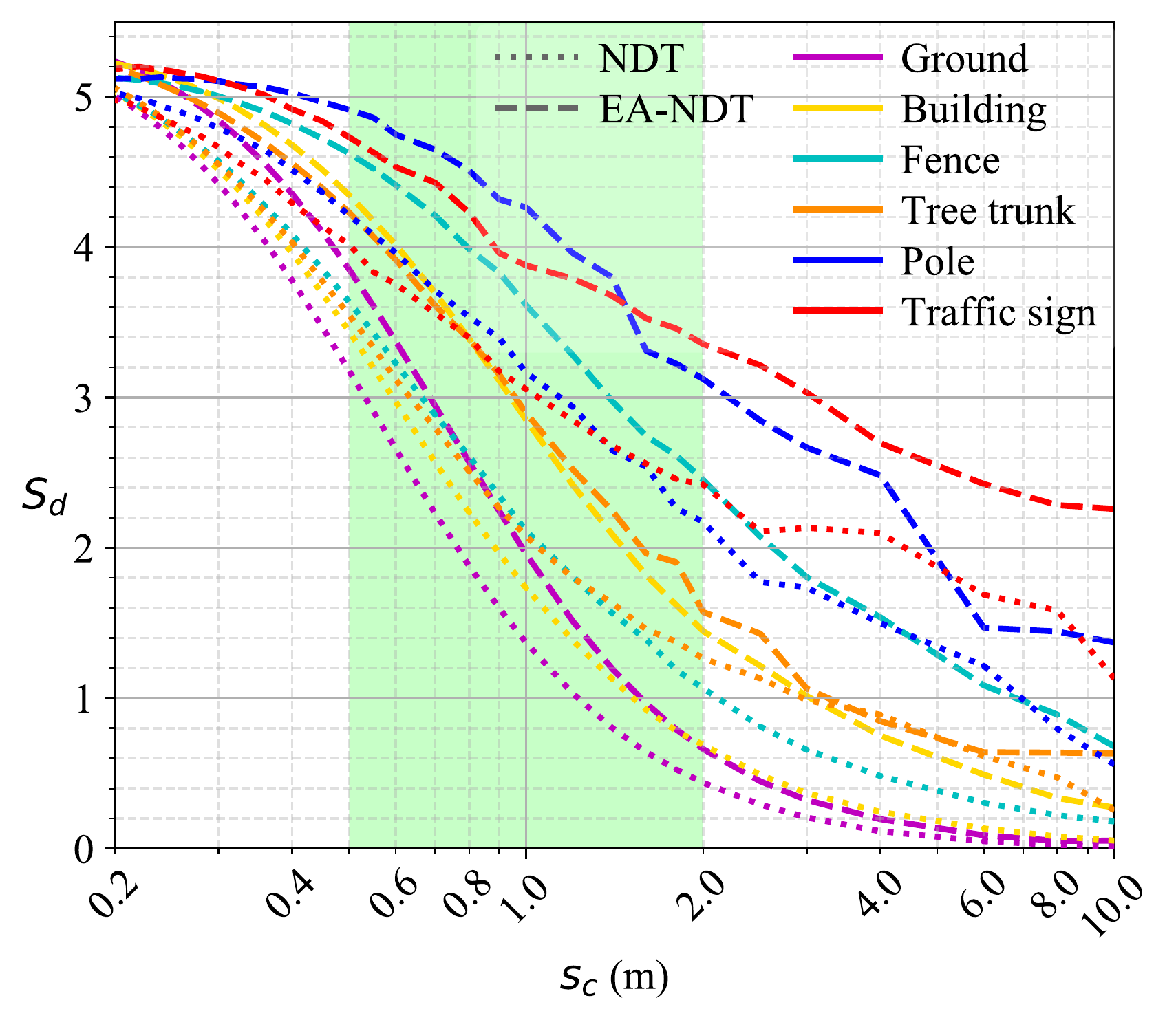}
    \caption{Comparison of descriptivity score $S_d$ of each label w.r.t. cell size $s_c$, line style indicates the method, color the label, and green background the applicable range.}
    \label{fig:label score vs resolution}
\end{figure}

\begin{figure}[!t]
    \centering
    \includegraphics[width=\linewidth]{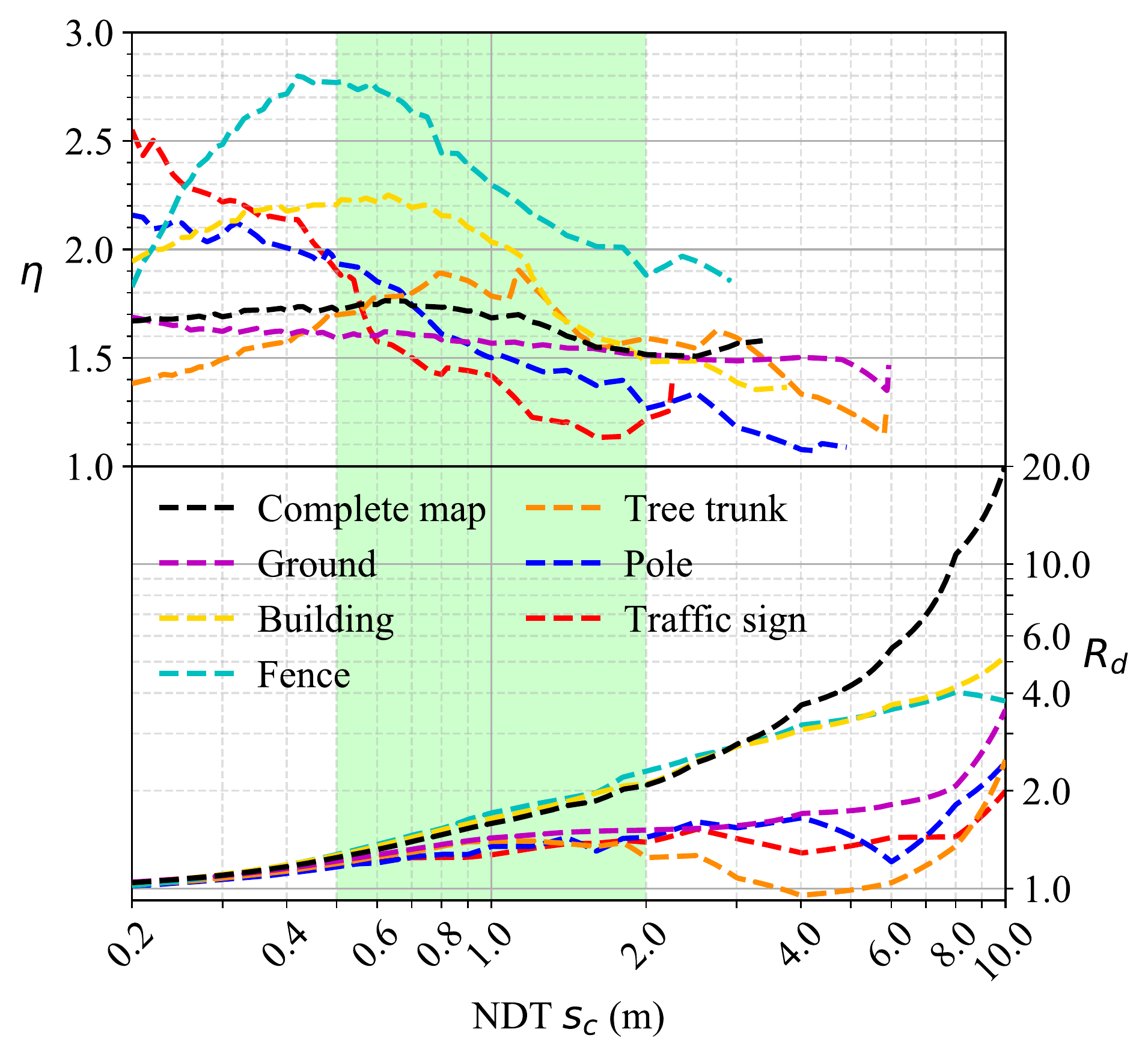}
    \caption{Both NDT compression efficiency $\eta$ (above) and descriptivity ratio $R_d$ (below) of the proposed method are visualized for the complete map and all labels w.r.t NDT cell size $s_c$, green background emphasizes the applicable range.}
    \label{fig:data compression ratio}
\end{figure}

In this section, we evaluate the quality between EA-NDT and NDT map representations and demonstrate the data compression of EA-NDT. The performance of both methods was evaluated by stepping the cell size from 0.2~m to 10~m with 30 values. Note that the computational time increases exponentially with decreasing cell size. The lower boundary of 0.2~m was selected since it could still be computed overnight. Similarly, the upper boundary of 10 m was considered large enough for this test. In Figs.~\ref{fig:complete score vs resolution}--\ref{fig:data compression ratio}, a practically applicable range of 0.5 -- 2.0 m, based on previous work \cite{magnusson2009three}, is highlighted. 


The evaluation of complete map representation on the dataset described in Section~\ref{subsec:experiments:dataset} is shown in Fig.~\ref{fig:complete score vs number of leafs}. It shows that EA-NDT map representation provides a higher descriptivity score for any number of cells (note that the minimum number of cells is limited for EA-NDT as explained in Section~\ref{subsec:experiments:implementation}). However, typically the results of NDT are compared as a function of the cell size, and therefore, in Fig.~\ref{fig:complete score vs resolution} we present an alternative comparison for which the number of cells in EA-NDT were fitted with NDT as explained in Section~\ref{subsec:experiments:implementation}. Likewise, descriptivity of EA-NDT outperforms NDT with any cell size. By comparing Fig.~\ref{fig:complete score vs number of leafs} and Fig.~\ref{fig:complete score vs resolution}, one can note that both plots are equally capable of showing the differences between the compared methods. In general, it can be noticed that the descriptivity score of NDT approaches EA-NDT with smaller cells. However, this is an expected phenomenon of grid cell division; the probability of multiple objects to be associated within one cell decreases with smaller cells.

In Table~\ref{table:dataset_characteristics} in Section~\ref{subsec:experiments:implementation}, it is shown that around 78.5 \% of the data consists of points labelled as ground or building, which reflects a similar proportion to the number of cells shown in Fig.~\ref{fig:fitting_leaves_and_resolutions}. The descriptivity score of the complete map is dominated by these abundant labels leaving the effect of other labels imperceptible. Therefore, in Fig.~\ref{fig:label score vs resolution}, we present an equivalent descriptivity score comparison separated for each label, which in case of NDT is equivalent to SE-NDT representation \cite{zaganidis2017semantic}. Similarly to the comparison of complete map representation, EA-NDT descriptivity score of each separate label is higher except for tree trunks with 3 -- 6 m cells, for which the descriptivity score equals with NDT. The low descriptivity of EA-NDT is most likely caused by the use of K-means clustering because if a cluster is large or a diameter of a trunk is small, a single cluster can contain points within the entire circumference of the trunk resulting in a non-Gaussian distribution. Moreover, the use of HDL-64 vertical FOV limits the height of tree trunks and poles in to a range of 2.5 -- 3 m (as explained in Section~\ref{subsec:experiments:dataset}) and when the required cell size exceeds half of that height, a large portion of the primitives is assigned into a one cluster instead of two causing the observed discontinuity. However, because of scaling the number of cells, the effect does not appear exactly with the expected cell size. With tree trunk and pole labels it is also observable that the descriptivity does not decrease with the largest cells, because the size of the primitive limits the cluster size from increasing.

The descriptivity ratio between EA-NDT and NDT is shown in the lower part of Fig.~\ref{fig:data compression ratio}. In general, the descriptivity ratio increases for all the labels towards the greater cell sizes, tree trunk and pole labels make an exception that was already covered above. Within the applicable cell range, the improvement in descriptivity ratio is more constant for all labels. For the complete map with 2 m cells, the descriptivity is 2$\times$ higher compared to NDT and for 10 m cells the descriptivity is 20$\times$ higher. Especially, building and fence labels show relatively higher descriptivity scores, which suggests that the plane extraction is advantageous.

Map compression is a direct consequence of EA-NDT's higher descriptivity scores; EA-NDT achieves the same descriptivity with a larger cell size which means a smaller number of cells. The NDT compression efficiency $\eta$, visualized in the upper part of Fig.~\ref{fig:data compression ratio}, was used to compare compression of EA-NDT with NDT (note that $\eta$, shown in Fig.~\ref{fig:complete score vs number of leafs}, can be computed only when a corresponding score exists for both methods). For the complete map representation, EA-NDT provides 1.5 -- 1.75$\times$ better compression within the whole examined range. The compression of the complete EA-NDT is mainly defined by the ground label, which is about 1.5$\times$ better within the whole range. EA-NDT's compression of traffic sign and pole labels is more than 2.1$\times$ higher than NDT for the smallest cells but drops steeply towards greater cell sizes, though, remaining higher compression even for the largest cells. For building and fence labels, the compression is more than 2.2$\times$ higher around 0.5 m cell size, for smaller and larger cells the NDT compression efficiency decreases. This suggests that EA-NDT's technique of modeling the planes and excluding the other points is beneficial until cell size of about 0.5 m but with smaller cells, NDT reaches the difference by modeling the excluded points. Finally, we can state that the complete map representation of EA-NDT achieves the highest compression improvement around 0.7 m cell size which is also within the applicable cell size range.

\section{Discussion} \label{sec:discussion}

The proposed EA-NDT achieves 1) at least 1.5$\times$ higher compression, and 2) always a higher descriptivity score with the same number of cells compared to NDT as shown in Fig.~\ref{fig:data compression ratio}. For separately tested semantic labels, EA-NDT achieves 1) always a higher compression, and 2) a higher descriptivity score in the applicable cell size range of 0.5 -- 2.0 m. However, we suggest to use cell sizes of 0.5 -- 1.0 m for EA-NDT in a suburban environment since our results (Fig.~\ref{fig:data compression ratio}) indicate that this range provides better compression.

Due to the semantic-aided instance clustering and primitive extraction, the proposed EA-NDT is able to find the most significant planar and cylindrical primitives in the environment. NDT representation is especially informative within planar and thin cylindrical structures that can be modeled with a unimodal distribution, and therefore, EA-NDT is able to model the environment more optimally compared to NDT. Moreover, the use of semantic information enables selection of the stable objects that should be modeled in the map. Finally, K-means clustering of the primitives ensures data efficient placement of cells where they are needed. The advantage of EA-NDT is a result of improved point cloud division. Therefore, the advantage is prominent in small objects such as poles or complicated structures such as buildings or fences. In the ground plane, the advantage is less evident because it is in any case a one large plane and the benefit can be almost completely explained by the removal of outliers and by the efficiency gained from clustering the ground plane.

Finding planar primitives in building and fence instances removes some points which are not modeled by EA-NDT HD Map. As shown in our results in Fig.~\ref{fig:data compression ratio}, that is beneficial for compression with cell sizes above 0.5 m but for smaller cell sizes it could be beneficial to model those points with additional NDT cells. However, as shown in this work, the described effect is not significant within the range of the suggested cell sizes. Moreover, the suggested correction should be justified only if it improves also the performance of scan registration.

The classification accuracy of the pre-trained RandLA-Net (see Section~\ref{subsec:experiments:dataset}) was a limiting factor for the quality of semantic segmentation. The misclassification increases the total number of cells when overlapping cells of different semantic labels model the same object (see Fig.~\ref{fig:visual abstract}), which reduces the compression of EA-NDT representation. In future work, the classification accuracy of the semantic segmentation could be improved by using a more advanced model \cite{tang2020searching, zhu2021cylindrical} and by retraining the model for the used LiDAR.

The proposed EA-NDT was tested in a suburban environment that consist of 1) a flat ground, buildings, fences, and traffic signs, which are modeled as planar surfaces, and 2) poles and tree trunks, which are modeled as cylindrical objects. The tested environment contains enough samples of all the tested semantic labels to demonstrate that EA-NDT is able to compress the data more than NDT. However, our tests did not concern vegetation, tree canopies, water, significant height variations, nor high rise buildings. In the future, a larger variety of environments should be studied. 

EA-NDT provides map compression within the tested semantic labels in environments where 1) the used semantic labels exist in the environment, 2) the reliability of semantic segmentation is high enough, and 3) instances are separated by sufficient distance. In order to use the proposed EA-NDT, the following assumptions need to hold: 1) ground, buildings, fences, and traffic signs must be composed of planar surfaces, and 2) tree trunks and poles need to be cylindrical. In future work, for other semantic labels, the type of primitive would need to be defined according to the properties of that label.

Semantic information is a powerful tool and a key enabler of HD Maps. In this work, we have shown that semantic information enables separate processes for each semantic label which results into more optimal clustering of point cloud data. Furthermore, in previous works semantic information has been used to improve positioning \cite{zaganidis2017semantic, zaganidis2018integrating, cho2020semantic}. Moreover, semantic information enables the removal of unwanted dynamic objects from the map. In future work, the use of semantic information opens a possibility to study the positioning accuracy and reliability of different object types over time. That could be especially useful when navigating in constantly changing environments such as arctic areas.

This work was outlined on evaluating compression and descriptivity properties of EA-NDT HD Map, and therefore, the positioning performance of the proposed framework remains an open question for the future work. Although, the positioning was not evaluated, the well established scan registration and cell representation of NDT is integrated into positioning of EA-NDT. Moreover, the data compression of an HD map is a desired property of any mobile robot application.
Another open question is how the proposed EA-NDT HD map can be efficiently updated with new information. 
Also, currently the computation of EA-NDT is very slow and the computational optimization is left for future work.

\addtolength{\textheight}{-0.121cm}   

\section{CONCLUSIONS} \label{sec:conclusion}

In this work, we proposed EA-NDT, that is a novel framework to compute a compressed map representation based on NDT formulation. The fundamental concept of EA-NDT is semantic-aided clustering to find planar and cylindrical primitive features of a point cloud to model them as planar or elongated normal distributions in a 3D space, respectively.

We showed that compared to NDT, the data-driven approach of EA-NDT achieves consistently at least 1.5$\times$ higher map descriptivity score, and therefore enables a significant map compression without deteriorating the descriptive capability of the map. The best compression in comparison to NDT is obtained within cell sizes of 0.5 -- 2 m, which is an applicable range for real-time positioning. Moreover, the results show that compared to NDT, the representation achieves a higher data compression within all the tested semantic labels, that is a desired property for mobile robots such as autonomous vehicles.

When data compression is a required property of an HD map, we recommend the use of EA-NDT instead of NDT. Based on the results of this work, it seems likely that the positioning accuracy using EA-NDT maps exceeds that of standard NDT maps of same size. However, this warrants future studies because there are several interacting factors such as potentially varying contribution of different semantic labels to the positioning accuracy.

\section*{ACKNOWLEDGMENT}

In addition, the authors would like to thank Paula Litkey and Eero Ahokas from FGI for data management and collection and Antero Kukko and Harri Kaartinen from FGI for assistance and advices. We would also like to thank Leo Pakola for participation in the research vehicle development.


\bibliographystyle{ieeetr}
\bibliography{article}

\end{document}